# DeePLT: Personalized Lighting Facilitates by Trajectory Prediction of Recognized Residents in the Smart Home

*Danial Safaei\*, Ali Sobhani, Ali Akbar Kiaei*

**Abstract**

In recent years, the intelligence of various parts of the home has become one of the essential features of any modern home. One of these parts is the intelligence lighting system that personalizes the light for each person. This paper proposes an intelligent system based on machine learning that personalizes lighting in the instant future location of a recognized user, inferred by trajectory prediction. Our proposed system consists of the following modules: (I) human detection to detect and localize the person in each given video frame, (II) face recognition to identify the detected person, (III) human tracking to track the person in the sequence of video frames and (IV) trajectory prediction to forecast the future location of the user in the environment using Inverse Reinforcement Learning. The proposed method provides a unique profile for each person, including specifications, face images, and custom lighting settings. This profile is used in the lighting adjustment process. Unlike other methods that consider constant lighting for every person, our system can apply each 'person's desired lighting in terms of color and light intensity without direct user intervention. Therefore, the lighting is adjusted with higher speed and better efficiency. In addition, the predicted trajectory path makes the proposed system apply the desired lighting, creating more pleasant and comfortable conditions for the home residents. In the experimental results, the system applied the desired lighting in an average time of 1.4 seconds from the moment of entry, as well as a performance of 22.1mAp in human detection, 95.12% accuracy in face recognition, 93.3% MDP in human tracking, and 10.80 MinADE$_{20}$, 18.55 MinFDE$_{20}$, 15.8 MinADE$_{5}$ and 30.50 MinFDE$_{5}$ in trajectory prediction.

## 1 Introduction

One of the most significant issues in a smart home is lighting control. Adjusting the lighting should be done to reduce energy consumption and increase electricity consumption efficiency. On the other hand, it satisfies each person's ideal and desired lighting conditions.

Artificial intelligence and machine vision can solve some problems in smart homes. Machine vision applications in smart homes offer many intelligent approaches to control different home areas to increase 'residents' comfort and convenience, increase efficiency, and reduce energy consumption and costs. In recent years, many studies have been conducted to identify human activities for different applications in smart home areas [1]. One of these areas is home lighting, where ideal and optimal lighting conditions should be achieved. However, using machine vision in the real world, lighting control is complex and has several challenges. Detecting and recognizing persons in various scene conditions (in terms of light intensity, angle, and quality of capturing, occlusion, environmental dynamics, etc.), applying appropriate and efficient light settings, understanding the 'user's desired lighting, and the cost of implementing an intelligent lighting system are examples of challenges. Deep learning and inverse reinforcement learning are the key factors in overcoming these challenges and achieving progress in the smart home.

The deep learning approach has successfully solved many complex problems. According to [2], a deep neural network can extract high-level representation in the deep layers, making it more suitable for complex activity recognition tasks. Hongqing and Chen [3] compare deep learning methods for human activity recognition in the smart

Home with traditional methods, such as the Hidden Markov Model (HMM) and naïve Bayes classifiers (NBC). Their experimental results show that the deep learning algorithm is an effective way to recognize human activities in the smart home. One of the essential advantages of deep learning 'models is that data features can be extracted automatically without the need for various time-consuming and laborious methods [4]. Also, the features are generated hierarchically from low to high [5]. Singh et al. [6] used deep learning techniques for activity recognition from raw sensory inputs in the smart home, compared CNN and LSTM deep learning models and showed the high performance of deep learning models. All the research we have mentioned proves that deep learning models help improve smart home 'systems' performance.

The lighting system that we propose here is based on machine vision methods, the deep-learning approach, and inverse reinforcement learning. Unlike the old systems, where lighting is manually adjusted, this system can intelligently adjust the light based on identifying people and their desired lighting. In summary, the proposed system operates as follows; the system first senses a 'person's presence in the environment by a PIR sensor and then applies the default lighting condition. The next step, by human detection and face recognition algorithms, identifies the person and then apply their desired lighting. Then the system tracks the person until they leave the environment. In the last step, the trajectory prediction module of our system uses the output of the tracking module to infer the future location of the user or users in the environment.

In the following sections, we first mention some related works in section 2, then describe the proposed 'system's components in both the hardware and software parts in section 3. In the software section, we describe the architecture of the neural networks and algorithms used for human detection, face recognition, human tracking, and human trajectory prediction. In section 4, we test the system and evaluate its performance. Finally, in section 5, we present the research results.

## 2 Related Works

A broad scope of applications in the smart home has led to much research in this field. The purpose of smart homes is to improve efficiency and increase comfort. The smart 'home's basic concept, as a standard relation system integrating different services in a home, was introduced in 1992 [7]. Berlo et al. [8] introduced automatic control as a definition of a smart home. In another definition, Alam et al. [9] defined a smart home as an application of universal computing, which enables users to provide the automated conscious concept as environmental intelligence, remote control, or home automation.

Home automation systems often learn the life pattern by processing data collected from sensors to better adapt to environmental conditions. Recognizing human activity is one of the main steps in designing and developing smart homes. Brdiczka et al. [10] recognized the human behavioral model from observations based on audio and video information in the smart home environment. They analyzed audio streams from a microphone with a speech activity tracker. In addition, they used 3D video tracking to detect people on the screen in different situations using hidden Markov models. Park et al. [11] collected input data from a smart home with many sensors, identified with an LSTM model and residual-RNN deep neural network, and categorized the 'home's human activities.

Much research has proved that machine learning algorithms effectively recognize human behavior in the smart home. To recognize the activities, some activities are defined as unknown actions due to the complexity of events or changes in conditions. Fakhredanesh and Roostaie [12], with an unsupervised method, automatically detected action changes without referencing tagged data and classifying the actions. In that study, they used the HOG feature and a silhouette-based framework to demonstrate action and detect changes in unsupervised behavior in a video sequence. Mao et al. [13] explained that machine learning algorithms build a model from training datasets to make decisions or predictions. The purpose of using machine learning algorithms in the smart home is to improve performance. Today, some research in the smart home field is conducted based on the deep learning approach. This approach learns non-parametric models of the inner structure of complex and huge data via various, and numerous neural processing layers called deep neural networks [14]. Othman and Aydin [15] combined an IoT-based system with machine vision to detect human activity for security goals. For this purpose, they used Raspberry Pi 3 and PIR sensor hardware to monitor and recognize humans. Their work takes the image from the scene when the PIR sensor detects the activity. The support vector machine (SVM) classification of the histogram of oriented gradients (HOG) diagram features identifies the human in the picture. Their image is sent to the smartphone using the Telegram application.

**Table 1** Comparison related works in terms of the proposed approach, automation quantity, hardware cost, and identification capability

| Research | Purpose | Proposed Approach | A.Q[1] | H.C[2] | I.C[3] |
|---|---|---|---|---|---|
| Brdiczka et al. [10] | Recognizing the behavioral human model | -Speech activity tracking<br>-3D video tracking<br>-Hidden Markov models | High | High | No |
| Park et al. [11] | Categorizing people activities | - LSTM model<br>-Residual-RNN | High | High | No |
| Fakhredanesh and Roostaie [12] | Automatically detecting action changes | -Silhouette-based framework<br>-HOG features | High | Low | No |
| Othman and Aydin [15] | Detecting the human for security goals | -Support Vector Machine (SVM)<br>-HOG features | Low | Low | No |
| Mehr and Polat [1] | Recognizing the human activities | -Convolutional Neural Network | High | Medium | No |
| Peng et al. [16] | Controlling home equipment | -Reinforcement learning method<br>-Adaptive decision model | High | High | No |
| Chandrakar et al. [17] | Smart lighting control | -NFC technology | Low | Low | Yes |
| Chun and Lee [18] | Smart lighting control based on human motion tracking | -Depth information collected independently by the infrared sensor and the thermal sensor | High | High | No |
| Adnan et al. [19] | Smart lighting system based on IoT and human activity | -Manually select an action from the activities listed in a software application | Low | Low | Yes |
| Patchava et al. [20] | Controlling home equipment | -Integrating cameras and motion sensors into a web application | Low | Low | No |
| **Ours** | -Personalize lighting<br>-Smart lighting control | -Deep human detection<br>-Deep face recognition<br>-Target tracking<br>-Target trajectory prediction | High | Low | Yes |

Notice that the HOG-based methods need lots of time to extract and train with the categorization model. Mehr and Polat [1] recognized human activities using 'CNN's impressive architecture as a deep learning method in the smart home. Their study was the first on the DML Smart Actions dataset using deep neural network models. This architecture may not be a sufficient model for each dataset, and its 'model's performance depends on the dataset extension. Peng et al. [16] extracted scene information from graphic data by the reinforcement learning method. An adaptive decision model based on deep learning was constructed. A simulated system was tested based on its ability to predict the best interval present in a day for lights and blinds in different homerooms to update the home 'equipment's operations to suit the 'user's behavior.

Smart lighting is one of the most significant smart home applications, which adjusts the lighting based on custom style. Some research has been developed with sensors, actuators, and wireless networks in smart lighting in recent years. Chandrakar et al. [17] used NFC technology to perform a smart lighting control method. The NFC tag information is read and sent to the microcontroller, where the particular light profile is selected from a user-defined list of light profiles. This system requires direct user action, and the user should always have the NFC tag to adjust its lighting. Chun and Lee [18] performed a smart lighting system based on human motion tracking that provides automatic lighting control. The depth information is collected independently by the infrared sensor and the thermal sensor. They used commercial depth or thermal cameras to track and estimate human behavior accurately. The main problem in their system is restrictions on the use of depth cameras such as Kinect and thermal cameras. So, the implementation of this system with regular cameras is not possible.

Many smart home systems are based on IoT technology, which controls the lighting system in a smart home with some device, such as a smartphone or tablet. Adnan et al. [19] developed a smart lighting system based on IoT and human activity. In their approach, the appropriate light intensity is determined for different daily activities, and then the hardware for the smart lighting system is developed. They designed an application for the lighting control system and adjusted human activity with an Android smartphone. Users must register on the app and then select an action

---

[1] Automation Quantity: Shows how much a proposed system is automated for its purpose.
[2] Hardware Cost: Includes the cost of the processing unit, sensors, cameras, etc.
[3] Identification Capability: Recognize each person with biometric features or account information.

From the activities list. The application sends the appropriate lighting with selected activities through a Bluetooth connection to the system. In the next step, the system measures the environmental brightness with the light sensor and then compares these two values; if the background 'light's intensity is higher or equal to the desired value, the lightning will not change. Their method directly interacts with the user and always requires manual control. Patchava et al. [20] presented a system for smart home automation using a Raspberry Pi based on machine vision and IoT. Their system works by integrating cameras and motion sensors into a web application. They aimed to control the equipment and parts of the home, like lighting through a smartphone or laptop, using Raspberry Pi and a web-based application. Their system could also detect a 'person's presence and report to the homeowner. Although this method seems attractive, it depends on the user, and decisions are made directly with the 'user's intervention.

In the following sections, we present our proposed method based on machine vision and inverse reinforcement technologies to apply the 'users' desired lighting.

## 3 Proposed System

In general, the proposed 'system's operation (Fig. 1) is that when a person enters a part of the home (such as a room), his/her presence is detected by a PIR sensor, then the system applies its default lighting condition. The camera system is activated in the next step, and the human detection algorithm is applied to the captured images, localizing the person. Secondly, face detection and recognition algorithms are given spatial information to identify the person. The system adjusts lighting conditions to the 'person's desire, which is already stored in their profile. In the third step, the system runs the tracking algorithm and tracks them as long as the person leaves the environment (such as a room). If the tracking system loses the person in the images, the system returns to the detection stage, and if it cannot detect any person in the environment, the lighting will be turned off. The main point about the user is the history of their path in the environment, which predicts whether the user is leaving or entering another part of the environment. Therefore, at the final stage, our 'system's trajectory prediction module forecasts the 'user's future location and, based on that prediction, personalizes the light in the new environment (such as another room the user is heading to), which system predicts that the user would be in. This module makes our system more efficient in satisfying the desire of the detected user.

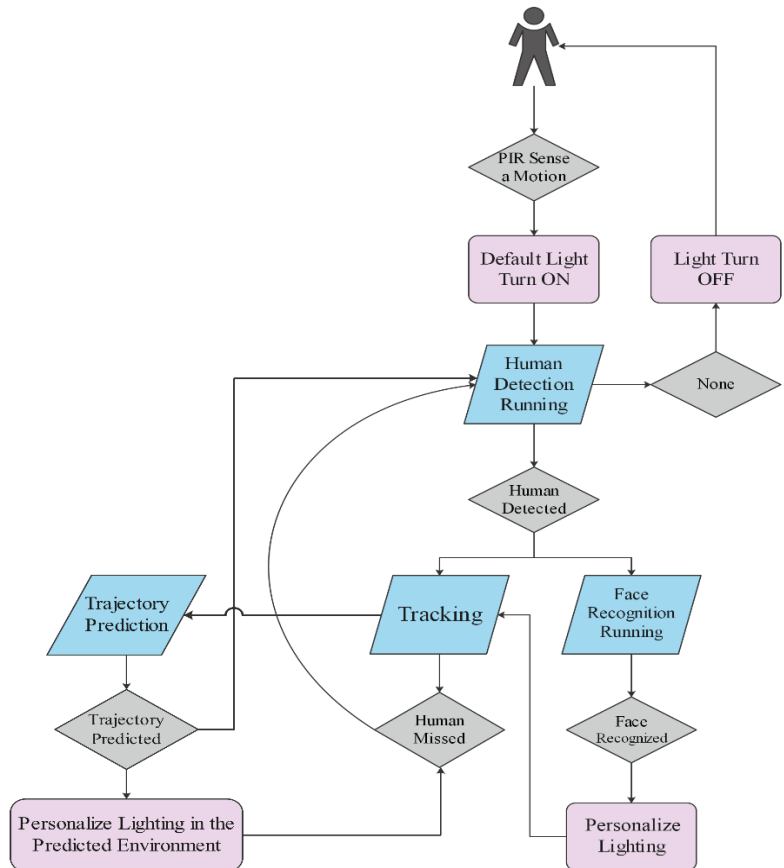

**Fig. 1** Flowchart of the proposed system operation: Blue: image processing algorithms, Pink: lighting changes, Gray: events

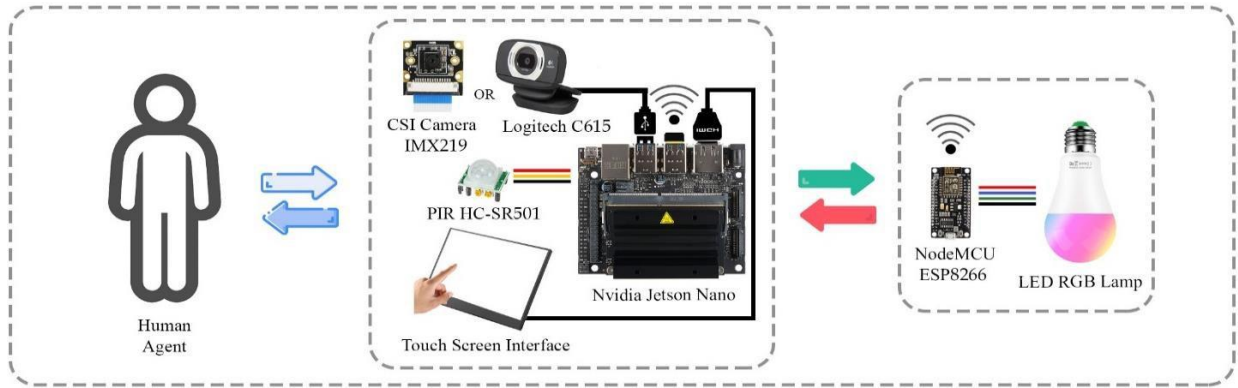

**Fig. 2** Hardware components of the proposed system: Central part (consisting of NVIDIA Jetson-Nano board, PIR sensor, camera, and touch screen interface), Wireless control and communication unit (NodeMCU), Lamps

Our proposed system consists of two general parts comprising hardware and software. The smart lighting system becomes operational by implementing the selected algorithms in the software part with the appropriate hardware.

### 3.1 Hardware Section

The hardware system we used in this study is demonstrated in Fig. 2. In the following: we introduce the main hardware sections.

#### 3.1.1 NVIDIA Jetson-Nano

Because we used deep neural networks in our proposed system, these networks contain many parameters and high computational load, so we needed a graphics processing unit (GPU) to increase the speed and accuracy performance of the network. The Jetson-Nano board has 128 CUDA cores, 472 GFLOPs, and 4 GB of RAM [21]. This module is much higher in processing power than similar boards like the Raspberry Pi series [22]. It has 40 GPIO pins, 4 USB ports, an HDMI port, and others. We used this board as the main hardware and its ports to connect to other hardware components.

#### 3.1.2 PIR Sensor

This sensor detects the environment's initial movement and turns on the lamps with default lighting conditions. In this study, we used the HC-SR501 sensor, which is compatible with the Jetson-Nano board and can be adjusted for sensitivity and timing.

#### 3.1.3 Camera

We needed a camera to capture images of the environment for detecting, recognizing, and tracking people. Here we can use a CSI camera or a webcam.

#### 3.1.4 Lighting Control System

There are two ways to adjust the brightness:
- Direct connection of the lighting circuit to the Jetson-Nano board and its direct lighting adjustment (wire connection).
- An indirect connection via an intermediate circuit that provides WIFI wireless connectivity. For example, using the NodeMCU module, commands can be received via a WIFI connection from the Jetson-Nano board, which adjusts the lighting. It is better to use RGB lamps using light adjustment, but it is possible to minimize the system requirement using different single-color lights.

#### 3.1.5 Touch Screen Interface

We needed a touch screen for registration, profile operation, and custom lighting adjustment to interact with the user.

## 3.2 Software Section

Our proposed approach in the software part consisted of models based on deep neural networks and image processing filters. This paper used models that provide optimal and optimized accuracy and efficiency in three stages of human detection, recognition, and tracking.

### 3.2.1 Human Detection

In this research, the MobileNetV2 algorithm in the SSDLite framework [23] was used to detect and determine the 'image's human position in real-time. The MobileNet neural network architecture has been designed specifically for limited hardware resources such as mobiles. With a dramatic decrease in the number of parameters and memory required, maintaining the same accuracy has become an advanced state-of-the-art model in computer vision for mobiles [23]. In the following section, its architecture is introduced.

**Table 2** Bottleneck residual block transforming from $k$ to $k'$ channels, with stride $s$, and expansion factor $t$ [23]

| Input | Operator | Output |
|---|---|---|
| $h \times w \times k$ | 1x1 conv2d, ReLU6 | $h \times w \times (tk)$ |
| $h \times w \times tk$ | 3x3 dwises=s, ReLU6 | $\frac{h}{s} \times \frac{w}{s} \times (tk)$ |
| $\frac{h}{s} \times \frac{w}{s} \times tk$ | linear 1x1 conv2d | $\frac{h}{s} \times \frac{w}{s} \times k'$ |

Table 2 Shows the MobileNetV2 architecture, which contains the initial fully convolution layer with 32 filters, followed by 19 residual bottleneck layers described in Table 3 [23]. In this architecture, the ReLU6 activation function is used as the non-linearity because of its robustness when used with low-precision computation [23], [24].

**Table 3** MobileNetV2: Each line describes a sequence of 1 or more identical layers, repeated $n$ times. All layers in the same sequence have the same number $c$ of output channels. The first layer of each sequence has a stride $s$, and all others use stride 1. All spatial convolutions use $3 \times 3$ kernels. The expansion factor $t$ is always applied to the input size described in Table 2 [23]

| Input | Operator | t | c | n | s |
|---|---|---|---|---|---|
| $224^2 \times 3$ | conv2d | - | 32 | 1 | 2 |
| $112^2 \times 32$ | bottleneck | 1 | 16 | 1 | 1 |
| $112^2 \times 16$ | bottleneck | 6 | 24 | 2 | 2 |
| $56^2 \times 24$ | bottleneck | 6 | 32 | 3 | 2 |
| $28^2 \times 32$ | bottleneck | 6 | 64 | 4 | 2 |
| $14^2 \times 64$ | bottleneck | 6 | 96 | 3 | 1 |
| $14^2 \times 96$ | bottleneck | 6 | 160 | 3 | 2 |
| $7^2 \times 160$ | bottleneck | 6 | 320 | 1 | 1 |
| $7^2 \times 320$ | conv2d 1x1 | - | 1280 | 1 | 1 |
| $7^2 \times 1280$ | avgpool 7x7 | - | - | 1 | - |
| $1 \times 1 \times 1280$ | **conv2d 1x1** | - | k | - | - |

The two types of blocks in the network are depicted in Fig. 3. One of the blocks has a residual connection with a stride of 1, while the other block does not have a residual connection and has a stride of 2. Each block comprises a point-wise convolution (1x1) layer, followed by a ReLU6 activation function, a depth-wise convolution (3x3 filter size) layer, another ReLU6 activation function, and another point-wise convolution (1x1) layer. The network also incorporates inverted residual connections that extend over the bottleneck layer and connect the thin blocks. [25].

In order to detect objects, a modified version of the Single Shot Detector (SSD) algorithm [26] is used with MobileNetV2 as a feature extractor [27] on the COCO dataset [28]. This algorithm is highly accurate and fast in detecting the position of people in captured images. The estimated position is then passed onto the face recognition and tracking sections. The human detection algorithm interacts continuously with the tracking section, and if the person is lost in the captured images, the detection algorithm is re-run to either locate the person or notify their absence.

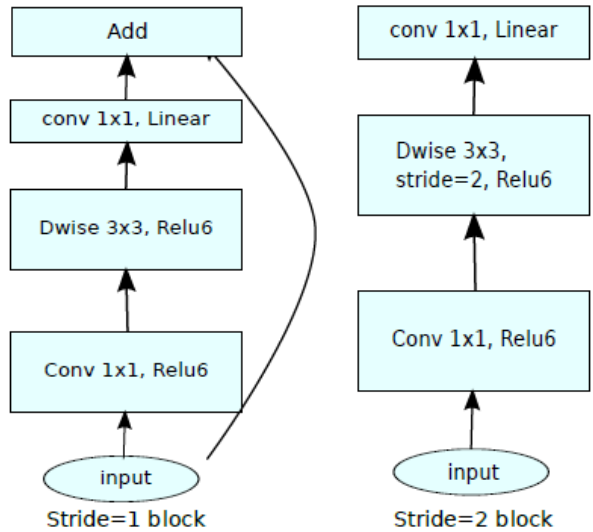

**Fig. 3** MobileNetV2 Block Architecture

### 3.2.2 Face Recognition

We need to identify the detected person in the captured image to adjust the lighting system according to their profile. In this section, accuracy in recognition and optimality is the essential criterion for selecting the algorithm. For this purpose, we use the FaceNet [29] algorithm to recognize the face. The FaceNet algorithm is based on a deep convolutional neural network and training a triplet loss, which directly fits a mapping of face images to a compact Euclidean space. The distances directly correspond to the criterion of face similarity. After producing this space, face recognition, verification, and clustering are implemented using standard techniques with embeddings from FaceNet as feature vectors [29]. More precisely, this model uses a small sample of images to produce the prototype, and when there are new models, the prototype can be used without re-training [30].

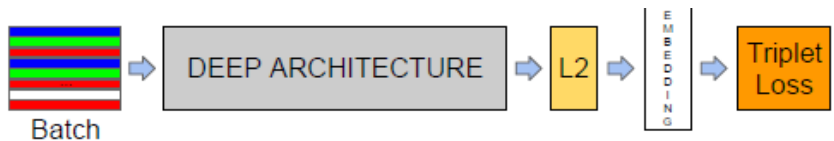

**Fig. 4** The model structure of the FaceNet: its network consists of a batch input layer and a deep CNN followed by L2 normalization, which results in face embedding. The triplet loss follows this during training [31]

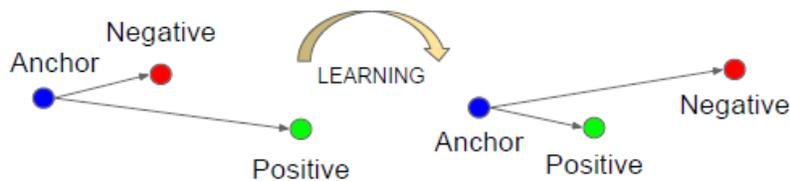

**Fig. 5** The triplet loss minimizes the distance between an anchor and a positive, both of which have the same identity, and maximizes the distance between the anchor and a negative of a different identity [31]

After recognizing the face by this algorithm, the 'person's desired lighting is applied. The identified 'person's ID label is assigned to the tracking section to be kept along with the 'person's location. If the person is unknown, the lighting will not change and remains in the default setting.

### 3.2.3. Tracking

The third step estimates the target position (human) in each image sequence frame in real-time. Since tracking is easier than detection, tracking algorithms can use fewer computational resources to detect objects per frame. Visual tracking requires strong filters to train a single frame and dynamically match the target appearance [32]. Human tracking in the environment requires control of changes on a large scale of complex image sequences. These changes are due to relative obstruction, deformation, motion blur, rapid motion, brightness changes, background changes, and scale changes.

Due to hardware limitations, the requirement for real-time image processing, and running multiple deep neural networks (to detect and identify people), we used a light algorithm based on a correlation filter introduced by Danelljan et al. [33].

The foundation of their work relies on the MOSSE tracker [32], a filter that introduced correlation filter technology to the visual tracking field through the Minimum Output Sum of Squared Error (MOSSE) filter [34]. With the ability to operate at 669 frames per second, this tracker algorithm is strong in detecting lighting, scale, pose, and non-rigid deformations variations [32]. While the MOSSE algorithm can efficiently adapt to small-scale variation, it lacks the ability to adjust to large-scale variation [34]. To address this limitation, Danelljan et al. proposed a discriminative correlation filter approach utilizing a scale pyramid representation for robust scale estimation. By implementing separate filters for translation and scale estimation, they demonstrated enhanced performance compared to an exhaustive scale search [33].

### 3.2.4. human Trajectory prediction

In the final step, as we mentioned before, the goal is to predict the user's future location in the environment. Before getting into details about how the trajectory prediction module works, we briefly review some preliminaries about maximum entropy inverse reinforcement learning (MaxEnt IRL) for path forecasting, conditioned on pre-defined goal states[35]–[37]. We could not only predict the agent's future trajectory using MaxEnt IRL in seen environments but also apply it for trajectory prediction in the unseen environment.

**MDP formulation:** For finite horizon settings with N steps Markov decision process is considered as $\mathcal{M} = \{\mathcal{S}, \mathcal{A}, \mathcal{T}, r\}$. $\mathcal{S}$ The state space consists of cells in a two-dimensional grid defined on the scene. $\mathcal{A}$ is the action space composed of 4 discrete actions, {up, down, left, right}, that could be used by the user to move to adjacent cells. $\mathcal{T} = \mathcal{S} \times \mathcal{A} \rightarrow \mathcal{S}$ is the state transition function that infers the next state of the user given the current state of the user and the action that the user takes. Finally, the reward function is defined by $r$. $r = \mathcal{S} \rightarrow \mathbb{R}_0^-$ is the reward function which maps each state to a real value less than or equal to 0. The initial state of the user and the goal state are defined by $s_{\text{init}}$ and $s_{\text{goal}}$ respectively. These two states of the MDP are known.

**MaxEnt IRL objective:** Under the condition of maximum entropy distribution, it is probable to observe a sequence of state actions $\tau = \{(s_1, a_1), (s_2, a_2), \ldots (s_N, a_N)\}$ commensurate with its exponential reward function.

$$P(\tau) = \frac{1}{Z} \exp(r(\tau)) = \frac{1}{Z} \exp\left(\sum_{i=1}^{N} r(s_i)\right) \quad (1)$$

In the equation, $Z$ is a normalizing constant. In maximum Entropy inverse reinforcement learning, the main objective is to retrieve a reward function $r_\theta(s)$ which is parametrized by a set of parameters $\theta$. This reward function takes the features of each state and returns a real value less than or equal to 0 for each of them. Using the reward function, which has been retrieved previously, the next step is to maximize the log-likelihood of observing a training set of demonstrations $T = \{\tau_1, \tau_2, \ldots, \tau_K\}$.

$$\operatorname*{argmax}_{\theta} \mathcal{L}_\theta = \operatorname*{argmax}_{\theta} \sum_{\tau \in T} \log\left(\frac{1}{Z_\theta} \exp(r_\theta(\tau))\right) \quad (2)$$

This equation could be solved using stochastic gradient descent. The gradient of log-likelihood $\mathcal{L}_\theta$ could be simplified to

$$\frac{d\mathcal{L}_\theta}{d\theta} = \sum_{\tau \in T} (D_T - D_\theta) \quad (3)$$

where $D_\tau$ and $D_\theta$ are state visitation frequencies (SVFs) for the training demonstrations $\tau$ and the expected state visitation frequencies for the Maximum Entropy policy given the current set of reward parameters $\theta$, respectively.

$\frac{d\mathcal{L}_\theta}{d\theta}$ could be obtained using backpropagation if we employ a deep neural network to model the reward function $r_\theta(s)$. In the following section, the algorithm which was employed to obtain $D_\theta$ is provided.

**Approximate value iteration:** Given the current reward function $r_\theta$ and the goal state $s_{goal}$ Algorithm 1 involves solving for the Maximum Entropy policy $\pi_\theta$. Given a state ""s"" the probability of taking action $a$ is being represented by $\pi_\theta$. The policy could be stationary or non-stationary. When the policy is fixed, it is independent of the time step. Stationary and non-stationary policies are indicated by $\pi_\theta(a|s)$ and $\pi_\theta^{(n)}(a|s)$ respectively. In our proposed method, we use a non-stationary policy. The algorithm involves iterative updates of the state and action log partition functions $V(s)$ and $Q(s,a)$. These functions can be interpreted as soft estimates of the expected future reward given state $s$ and the expected future reward given state-action pair *(s, a)*, respectively. In the initialization step $V(s)$ is initialized to 0 for goal states $s_{goal}$ And $-\infty$ for all other states. $V(s)$ and $Q(s,a)$ are then iteratively updated over $N$ steps while holding $V(s_{goal})$ fixed to 0. For each step $\pi_\theta$ is given by

$$\pi_\theta^{(n)}(a|s) = exp\left(Q^{(n)}(s,a) - V^{(n)}(s)\right) \quad (4)$$

The main reason why we hold $V(s_{goal})$ to 0 and all other $V(s)$ to $-\infty$ is to ensure that MDP ends at $s_{goal}$.

**Policy propagation:** In algorithm 2, state visitation frequencies are calculated. Starting with the initial state distribution algorithm two involves applying $\pi_\theta$ for $N$ steps repeatedly to give state visitation frequency at each stage. At each step, the state visitation frequency corresponding to the goal state is set to 0 since it absorbs any probability mass that reaches it.

---

**Algorithm 1 Approx. Value iteration (goal conditioned)**

Inputs: $r_\theta, s_{goal}$
1: $V^{(n)}(s) \leftarrow -\infty, \forall s \in S$
2: for $n = N, ..., 2, 1$ do
3:     $V^{(n)}(s_{goal}) \leftarrow 0$
4:     $Q^{(n)}(s,a) = r_\theta(s) + V^{(n)}(s'), s' = \mathcal{T}(s',a)$
5:     $V^{(n-1)}(s) = \log \sum_a exp(Q^{(n)}(s,a))$
6:     $\pi_\theta^{(n)}(a|s) = exp\left(Q^{(n)}(s,a) - V^{(n)}(s)\right)$
7: end for

---

**Algorithm 2 Policy propagation (goal conditioned)**

Inputs: $\pi_\theta, s_{init}, s_{goal}$
1: $D^{(1)}(s) \leftarrow 0, \forall s \in S$
2: $D^{(1)}(s_{init}) \leftarrow 1$
3: for $n = 1, 2, ..., N$ do
4:     $D^{(n)}(s_{goal}) \leftarrow 0$
5:     $D^{(n+1)}(s) = \sum_{s',a} \pi_\theta^{(n)}(a|s') D^{(n)}(s'), \mathcal{T}(s',a) = s$
6: end for
7: $D(s) = \sum_n D^{(n)}(s)$

---

**Path forecasting conditioned on goals:** To give path forecasts on a 2-D grid from the $s_{init}$ to $s_{goal}$, the maximum entropy policy $\pi_\theta^*$, for the converged reward model $r_\theta$, can be sampled from. The policy can explore multiple paths within the scene to the goal state because the maximum entropy policy $\pi_\theta^*$ Is stochastic. These explored paths would be sampled output trajectories.

## 4 Experimental Results

To evaluate the proposed 'system's performance, we first tested each software component of the system and then, in a field experiment, evaluated the 'system's overall performance with all software and hardware components.

### 4.1. Human Detection

The first step of system processing was to detect the 'person's position in the captured image, done by the MobileNetV2 [23] algorithm. Table 4 compares the performance of real-time object detection algorithms on the *COCO* [38] dataset. According to the results, the MobileNetV2+SSDLite achieved competitive accuracy with significantly fewer parameters and smaller computational complexity. This algorithm was not only the most efficient model but also the most accurate of the three. It was 20x more efficient and 10x smaller while still outperforming YOLOv2 [39] on the COCO dataset [23].

**Table 4** Performance comparison of MobileNetV2+SSDLite and other real-time detectors on the *COCO* [38] dataset object detection task [23]

| *Network* | *mAP* | *Params* | *MAdd* | *CPU* |
|---|---|---|---|---|
| SSD300[40] | 23.2 | 36.1M | 35.2B | - |
| SSD512[40] | 26.8 | 36.1M | 99.5B | - |
| YOLOv2[39] | 21.6 | 50.7M | 17.5B | - |
| MobileNetV1+SSDLite | 22.2 | 5.1M | 1.3B | **270ms** |
| MobileNetV2+SSDLite | 22.1 | 4.3M | 0.8B | **200ms** |

### 4.2. Face recognition

The second step of system processing is recognizing the detected 'person's face, done by the FaceNet [29] algorithm. To recognize the face, we needed to find the 'face's exact position in the 'person's extracted image. Therefore, we used the MTCNN [41] algorithm in this study. After determining the 'face's position, its location information was given to the FaceNet [29] algorithm to perform the recognition process.

The FaceNet [29] 'model's performance for face recognition achieved a 99.63% accuracy on the *Labeled Faces in the Wild (LFW)* [42] dataset and 95.12% accuracy on the *YouTube Faces DB* [43]. This system also cut the error rate compared to the best-published result [44] by 30% on both datasets.

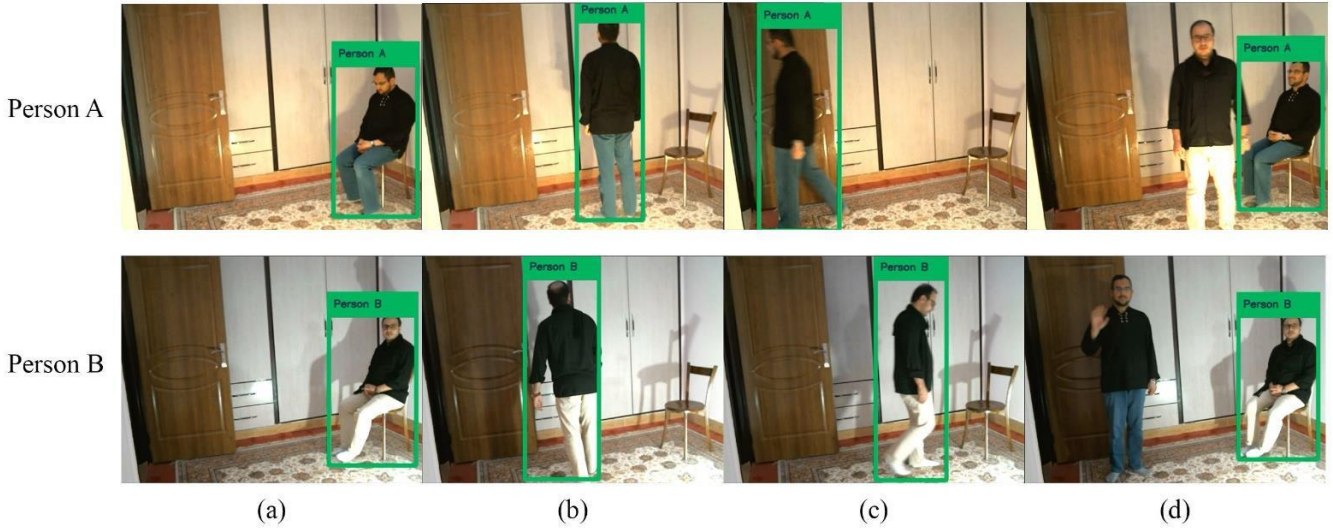

**Fig. 6** Field experiment of the proposed system in different situations: first row: custom lighting according to person A profile (yellow light), second row: custom lighting according to person B's profile (white light). Different scenarios; (a) sitting, (b) behind the camera, (c) walking, (d) the presence of different people in the environment with the priority of the first 'person's presence.

### 4.3. Person Tracking

The third step of the proposed system was tracking the detected person, which was implemented by a correlation-based filter algorithm introduced by Danelljan et al. [33]. The main reason for choosing this algorithm was its low computational cost and high processing speed, making it suitable for real-time tracking on limited hardware platforms such as mobile devices.

**Table 5** Precision test result of the proposed tracking algorithm in Danelljan et al. [33]

| Measure | Result (%) |
|---|---|
| **Median Overlap-Precision** | 75.5 |
| **Median Distance-Precision** | 93.3 |
| **Centre-Location-Error** | 10.9 |

The precision test result of this algorithm is shown in Table 5. This test was performed on the 28 benchmark sequences [45]. This result verified the ability of the algorithm to track objects with significant accuracy in overlap precision (OP) (%), distance precision (DP) (%), center location error (CLE), and with a high frame rate.\

### 4.4. Trajectory Prediction

In this section, we will outline our dataset collection methodology, introduce the metrics used to evaluate our system, and provide qualitative examples of our proposed system from the experimental results.

#### 4.4.1. Dataset

The human trajectory dataset we used in our work was collected in a 200-square-meter house involving two residents. The house map is shown in Fig. 7. A total number of 1000 trajectories are included for the dataset, with an average length of 5-10 meters long, in about 1000 meters of human trajectory. The input to our model would be snippets of agents' past trajectories and birds' eye view representations of the house.

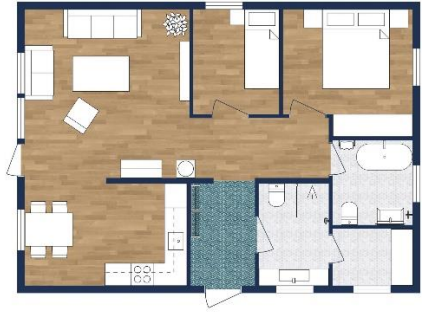

Fig. 7. Map of the house

### 4.4.2. Metrics

In order to assess the performance of our proposed model, we require a metric to gauge the extent to which its forecasts diverge from the actual future trajectory. Since our model produces forecasts from a multimodal distribution, we need a metric that is capable of accommodating trajectories generated by the model that may not precisely align with the ground truth. Thus, we will utilize the minimum K average displacement error (MinADE) and final displacement error (MinFDE) as our chosen metrics. These metrics have been used in prior works of multimodal trajectory forecasting [46]–[52]. Our model reaches 10.80 $MinADE_{20}$, 18.55 $MinFDE_{20}$, 15.8 $MinADE_5$ and 30.50 $MinFDE_5$ in trajectory prediction.

**Table 6** test results for the proposed trajectory prediction algorithm.

| Measure | Result (%) |
|---|---|
| $MinADE_{20}$ | 10.80 |
| $MinFDE_{20}$ | 18.55 |
| $MinADE_5$ | 15.8 |
| $MinFDE_5$ | 30.50 |

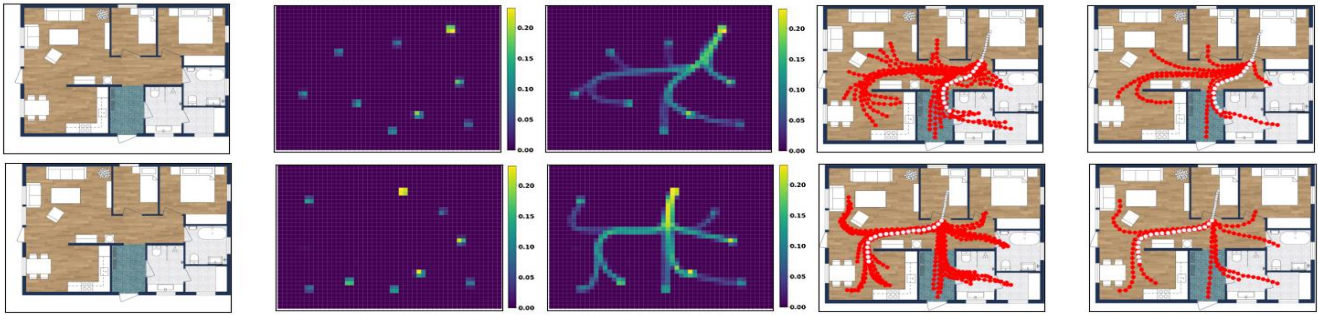

Fig. 8. Qualitative examples

### 4.4.3. Qualitative examples

Fig. 8. illustrates qualitative examples of our proposed module. From left to right first column is the input to our module, which is the history of user trajectory and a birds-eye-view map of the house. The left middle indicates the goals of the agent, which Are different parts of the house. The middle shows the path to these goals by modeling the agent as a MaxEnt policy exploring the house. The middle right shows the trajectories sampled from the policy. Sampling itself can be inefficient; for example, several sampled state sequences and trajectories could be identical or similar. This is because the MaxEnt policy induces a multimodal distribution over path and goal states. To provide downstream, we cluster the sampled trajectories using the K-means algorithm to output a set of K-predicted trajectories. Finally, the module outputs k predicted trajectories by clustering the sampled trajectories.

### 4.5. Field Experiment of the Proposed System

We first implemented the proposed hardware platform in a room to test the overall system performance. Then we created two profiles named Person A and Person B with different lighting adjustments on the system. To test the performance, we considered different scenarios for the entry and exit of persons, which are visible in Fig. 6.

**Table 7** Video specifications in the field experiment

| *Property* | *Unit* | *Value* |
|---|---|---|
| **Video-Resolution** | **Pixel** | **800 × 600** |
| **Frame-Rate** | **FPS** | **24** |
| **Duration** | **Second** | **293** |
| **Total-Frame** | **-** | **7032** |

The imaging system was performed by a camera attached to the Jetson-Nano board, and the specifications of the recorded video for processing are listed in Table 6. The speed of the detection, recognition, and tracking processes on the proposed hardware is given in Table 7. Duration in each-Episode is the interval between the arrival of the person and the application of his or her desired lighting. As demonstrated in this table, the most time-consuming process is the face recognition phase, with an average speed of 58 milliseconds per frame performed by the FaceNet [31] algorithm. Also, the tracking process is about 2.6x faster than the detection process performed by the MobileNetV2+SSDLite [53] algorithm.

**Table 8** The performance speed of algorithms on Jetson-Nano hardware and board

| Measure (on average) | Unit | Val |
|---|---|---|
| Human-Detection Rate | Millisec | 45 |
| Face-Recognition Rate | Millisec | 58 |
| Tracking Rate | Millisec | 17 |
| Trajectory Prediction | Millisec | 79 |
| Duration in each-Episode | Second | 1.4 |

The time interval between the 'person's arrival in the environment and applying the desired lighting includes the duration of the following processes: human detection, face recognition, finding the 'person's profile information in the database, sending information, and light adjustment commands to the lighting control circuit. Here, this duration is called an *Episode*, which takes about 1.4 seconds to complete. It should be noted that this time can be much more or less, based on conditions such as capturing angle, entry style, environmental characteristics, etc.

## 5 Conclusion

We proposed a system to personalize lighting intelligently. This system consists of machine learning algorithms such as human detection, face recognition, human tracking, and trajectory prediction implemented by the proposed hardware. Our system achieved better automation quantity, speed, and accuracy performance in personalized lighting than traditional lighting methods performed by switches, sensors, and even home management applications. Unlike many methods, this system is completely intelligent and does not require direct user intervention and control. The desired lighting is adjusted from the moment a person enters, and as long as the user stays in the home, our proposed system personalizes lighting wherever it predicts the user will head. The lighting does not change until the person leaves the environment, even if other people enter. The 'system's performance in detection, recognition, tracking, and trajectory prediction processes achieved acceptable results in the field experiment and satisfied the desired lighting requirements in a significantly short time.

In the proposed system, tracking people through deep learning methods will perform the process more accurately. Also, using an activity recognition algorithm for customizing lighting based on each 'person's activity will improve the 'system's performance. Due to the proposed 'system's hardware limitations in running multiple algorithms simultaneously, using these approaches is challenging. Using more powerful hardware, cloud computing, or hardware accelerators such as the *coral-USB* [54] accelerator is suggested for future work.

As an essential option for future work, we can integrate a reinforcement learning module in our proposed system to learn each home resident's desired lighting based on their long-term behavior. This new module learns different policies for residents, which could set the desired lighting for each user based on their behavior. This module would act adaptively, which means that we could learn multiple policies to satisfy various lighting demands of different users by using reinforcement learning. For instance, some people are more into warm colors, but others may be more into cold colors. Based on the uniquely generated policies, different color demand from different users could be satisfied.

## 6 Acknowledgments

By researchers in Machine Learning, Computer Vision and Deep Learning.

## 7 Funding

## 8 Conflicts of interest

## 9 Data availability